\documentclass[runningheads]{llncs}
\usepackage[T1]{fontenc}
\usepackage[utf8]{inputenc}
\usepackage{CJKutf8}
\usepackage{graphicx}%
\usepackage{multirow}%

\usepackage{amsmath,amssymb,amsfonts}%
\usepackage{amsthm}%
\usepackage{mathrsfs}%
\usepackage[title]{appendix}%
\usepackage{xcolor}%
\usepackage{textcomp}%
\usepackage{manyfoot}%
\usepackage{booktabs}%
\usepackage{algorithm}%
\usepackage{algorithmicx}%
\usepackage{algpseudocode}%
\usepackage{listings}%
\usepackage[colorlinks=true,
            linkcolor=blue,
            citecolor=blue,
            urlcolor=black]{hyperref}
\usepackage{url} 
\usepackage{cite}%
            
\begin{document}
\title{Temperature Field Reconstruction of Tungsten Monoblock Divertor on EAST using Physics-aware Neural Operator Transformer}

%
%
\author{
\href{mailto:e24201056@stu.ahu.edu.cn}{Zikang Yan} \inst{1} \and 
\href{mailto:xiaowang@ahu.edu.cn}{Xiao Wang}\inst{1}\thanks{Corresponding author: Xiao Wang, Zhendong Yang} \and  
\href{mailto:yangqq@ipp.ac.cn}{Qingquan Yang} \inst{2} \and 
\href{mailto:dongyz@tlu.edu.cn}{Zhendong Yang} \inst{3}$^\star$ \and 
\href{mailto:gaoting.chen@ipp.ac.cn}{Gaoting Chen} \inst{2, 4} \and 
\href{mailto:zhc23thuml@tsinghua.edu.cn}{Zehua Chen} \inst{5} \and 
\href{mailto:jiangbo@ahu.edu.cn}{Bo Jiang} \inst{1} \and 
\href{mailto:tangjin@ahu.edu.cn}{Jin Tang} \inst{1} \and 
\href{mailto:gsxu@ipp.ac.cn}{Guosheng Xu} \inst{2} 
}    
\authorrunning{Zikang Yan et al.}
%

\institute{
School of Computer Science and Technology, Anhui University, Hefei, China \and  
Institute of Plasma Physics, Chinese Academy of Sciences, Hefei, China \and 
Tongling University, Tongling 244000, China \and 
College of Physics, Donghua University, Shanghai 201620, China \and 
Tsinghua University, Beijing, China 
}

\maketitle              

\begin{abstract}
Accurate modeling of the divertor temperature field is essential for preventing material melting and damage and for extending the service life of fusion devices. However, conventional numerical methods, such as the Finite Element Method (FEM), are computationally expensive and therefore unsuitable for real-time applications. Therefore, a fast and generalizable method is required for real-time reconstruction of the divertor temperature field and subsequent real-time control. To address the above issue, we propose a Physics-aware Neural Operator Transformer (PNOT) to characterize the spatiotemporal evolution of the divertor temperature field. It models boundary heat-flux relations as a structured graph and employs graph attention to explicitly capture spatial physical dependencies. Inspired by physics-aware attention, we further develop a physics-aware neural operator module to aggregate query points with similar physical conditions via slicing and model heat diffusion, while a gradient-constrained Sobolev regularization loss enforces consistency between function values and their derivatives.
Experimental results show that these physical constraints improve prediction accuracy while preserving physical consistency. 
The source code of this paper will be released on \url{https://github.com/Event-AHU/OpenFusion} 

\end{abstract}



\section{Introduction}

\begin{figure}
\centering
\includegraphics[width=\textwidth]{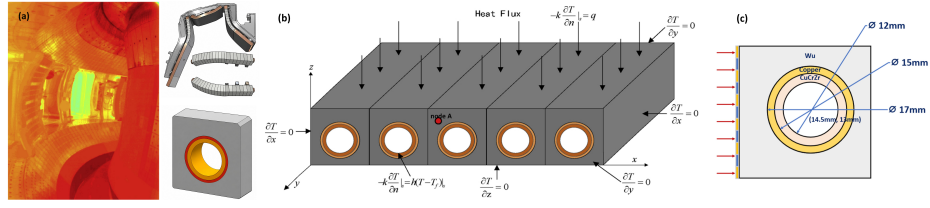} 
\caption{Divertor system of the EAST device. (a) Infrared-view image of the interior of the EAST device and schematic of the divertor model; (b) Physical model of the EAST divertor and the corresponding boundary conditions; (c) Cross-sectional modeling configuration of a single divertor monoblock.}
\label{fig::background}
\end{figure}

Controlled Nuclear Fusion (CNF) is widely regarded as one of the most promising solutions to future energy challenges. By fusing light atomic nuclei under extremely high-temperature and high-pressure conditions, fusion releases enormous amounts of energy through the same mechanism that powers the Sun, earning fusion devices the name \textit{artificial suns}. State-of-the-art fusion facilities worldwide consist of DIII-D~\footnote{\url{https://d3dfusion.org/}}, EAST (Experimental Advanced Superconducting Tokamak)~\footnote{\url{http://east.ipp.ac.cn/}}, HL-3~\footnote{\url{https://en.wikipedia.org/wiki/HL-2M}}, and others.
Due to the complexity of fusion systems and the strongly coupled plasma processes involved, a variety of diagnostic systems have been developed to monitor plasma conditions and obtain key physical parameters for experimental analysis and operational control. In plasma-facing components, the divertor plays a crucial role as it exhausts heat and impurities from the plasma while protecting the first wall from excessive thermal loads, as shown in Fig.~\ref{fig::background} (a), which illustrates the infrared view of the EAST device interior as well as the divertor module and a single divertor target.
To prevent damage or even melting of the divertor target plate under high heat loads, modeling its temperature field and performing heat flux analysis based on the temperature distribution are of great importance for the operational control of the EAST device.
As shown in Fig.~\ref{fig::background} (b), the schematic of the divertor target plate is presented. The top surface is directly exposed to plasma heat flux, while the lateral boundaries are assumed to be adiabatic. The central region consists of a water-cooling structure that continuously removes heat. Fig.~\ref{fig::background} (c) illustrates the cross-sectional structure, where the material configuration from the outer layer to the inner layer is composed of W (tungsten), Cu (copper), and CuCrZr.

Traditional numerical methods, such as the Finite Element Method (FEM), can accurately solve complex physical problems but require repeated discretization and iterative computation, making them unsuitable for real-time control and online prediction in fusion devices.
Recent advances in artificial intelligence, particularly deep neural networks~\cite{wang2024mambasurvey} and large foundation models~\cite{wang2023llmsurvey}, have driven the adoption of machine learning techniques in fusion research. Among them, Physics-Informed Neural Networks (PINNs)~\cite{wang2025heatflux, karniadakis2021physics} and Neural Operators~\cite{jha2025NeuOperatorSurvey} have emerged as promising approaches for solving complex physical problems. PINNs embed governing equations and physical constraints into the loss function, enabling learning with limited or even no labeled data. 
Neural Operators learn mappings between function spaces and exhibit stronger generalization across different parameters and scenarios. These advantages make Neural Operators particularly attractive for fast prediction and real-time applications in fusion systems.

As a fundamental architecture of modern foundation models, the Transformer~\cite{vaswani2017Transformer} has achieved remarkable success in natural language processing, computer vision, and scientific machine learning due to its strong capability for modeling long-range dependencies. Recently, Transformer-based neural operators have also demonstrated promising performance in solving Partial Differential Equations (PDEs).
GNOT~\cite{hao2023gnot} formulates operator learning as a generalized attention problem and achieves strong generalization across irregular domains.
Transolver~\cite{wu2024transolver} introduces a physics-inspired latent slice attention mechanism to improve computational efficiency while preserving global physical dependencies.
To further enhance scalability, DPOT~\cite{hao2024dpot} adopts large-scale autoregressive pretraining on diverse PDE datasets, significantly improving transferability across downstream physical systems. More recently, GAOT~\cite{wen2026geometry} incorporates geometry-aware attention to better model PDEs defined on arbitrary computational domains.
These advances demonstrate the effectiveness of Transformer-based neural operators for scientific computing and motivate their application to heat conduction problems. 
However, their application to heat conduction problems still faces several challenges: 
(1). Existing approaches~\cite{lu2021learning,rahman2022u,hao2023gnot,hao2024dpot} typically compress boundary conditions into a single global feature vector, potentially obscuring the spatial distribution and functional structure of boundary heat fluxes. 
(2). Standard self-attention mechanisms primarily capture global interactions and lack explicit modeling of local spatial coupling during heat diffusion, limiting the incorporation of physical priors. 
(3). Conventional $L_2$ loss functions supervise only solution values while neglecting gradient information, resulting in suboptimal physical consistency and generalization ability.

To address these limitations, we propose \textbf{PNOT}, a Physics-aware Neural Operator Transformer framework for heat conduction equations. Specifically, PNOT introduces a structured boundary representation to preserve spatial boundary information, incorporates a physics-aware feature interaction mechanism together with a spatial heat diffusion propagation module to encode local heat transfer priors, and adopts a Sobolev-constrained optimization strategy that jointly supervises solution values and gradients. An overview of our proposed PNOT framework is illustrated in Fig.~\ref{fig::framework}.
Extensive experiments on the finite-element dataset of the EAST divertor demonstrate that PNOT consistently outperforms state-of-the-art neural operator models in temperature field reconstruction. In particular, the proposed framework achieves lower reconstruction errors under diverse heat-flux conditions, preserves local temperature gradients more faithfully, and exhibits stronger out-of-distribution generalization across unseen operating conditions. Ablation studies further verify the effectiveness of each proposed component, while qualitative results demonstrate physically consistent and spatially smooth temperature field reconstruction.

In summary, the main contributions of this work are as follows:

(1) We propose a Physics-aware Neural Operator Transformer (PNOT) framework that incorporates a \emph{Flux Enhancement} module, a \emph{Physics Enhancement} module, and a \emph{Sobolev loss}, thereby enhancing boundary correlations, query interactions, and temperature gradient consistency for accurate temperature field reconstruction.

(2) We propose a \emph{Heat Graph Propagation} module and integrate it with \emph{Physics Attention} to explicitly model local and global spatial coupling and heat diffusion dynamics, thereby incorporating physically meaningful information into feature interactions. In addition, we further introduce a Sobolev-based training objective that jointly constrains solution values and spatial gradients, improving physical fidelity, predictive accuracy, and out-of-distribution generalization.

(3) Extensive experiments on the EAST divertor finite-element dataset show PNOT outperforms state-of-the-art neural operators in temperature field reconstruction, with lower errors, precise gradient preservation, and strong out-of-distribution generalization. Ablation and qualitative analyses verify the validity of our modules and the physical consistency of reconstructed thermal fields.

\section{Related Works}

\subsection{Neural Operator}
Neural operators have recently emerged as a promising paradigm for learning mappings between infinite-dimensional function spaces and have demonstrated remarkable success in solving parametric partial differential equations.
Neural operators have evolved from theoretical operator approximation to efficient and general-purpose architectures for scientific computing. Early work, represented by DeepONet~\cite{lu2021learning}, established the foundation of operator learning by directly learning mappings between functions. Later, the Fourier Neural Operator (FNO)~\cite{li2020fourier}significantly improved scalability and efficiency through spectral convolutions, inspiring several variants, including the Graph Neural Operator (GNO)~\cite{li2020neural} for irregular meshes, the Low-Rank Neural Operator (LNO)~\cite{kovachki2023neural} for computationally efficient kernel approximation, and the Wavelet Neural Operator (WNO)~\cite{tripura2023wavelet} for capturing localized and multi-resolution features. Subsequent developments incorporated geometric and physical priors through methods such as Geo-FNO~\cite{li2023geometry} and Physics-Informed Neural Operators (PINO)~\cite{li2024physics}. More recently, transformer-based approaches, including the Galerkin Transformer, have further enhanced the ability of neural operators to model complex systems, driving their evolution toward foundation-model paradigms for scientific machine learning~\cite{wu2024transolver,li2022transformer,hao2023gnot,cao2021choose}.
In recent years, neural operator pretraining has emerged as an important research direction in scientific machine learning. MoE-POT~\cite{wang2026mixture} and NESTOR~\cite{sun2026nestor} introduce the Mixture-of-Experts (MoE) mechanism into neural operators, leveraging dynamic expert routing to enhance model generalization across diverse partial differential equations (PDEs) and complex physical processes. DPOT~\cite{hao2023gnot} adopts an autoregressive denoising pretraining strategy to learn universal spatiotemporal evolution patterns from large-scale PDE datasets, thereby improving long-term forecasting performance. Latent Neural Operator Pretraining~\cite{wang2024latent}, on the other hand, employs latent-space modeling to compress physical field representations, reducing computational costs while improving the solution accuracy of time-dependent PDEs. Overall, these models have significantly advanced the application of artificial intelligence in the physical sciences, opening new avenues for data-driven PDE solving and the development of foundation models for physics.

\subsection{Temperature Field Reconstruction}
Heat flux analysis of divertors is a critical issue in fusion device research, and its essence lies in solving the heat conduction partial differential equations (PDEs). Traditional approaches mainly rely on numerical methods such as the finite difference method (FDM)~\cite{causon2010introductory}, finite element method (FEM)~\cite{reddy2026introduction}, and finite volume method (FVM)~\cite{eymard2000finite}. Among these, Shi et al.~\cite{shi2017heat} employed the two-dimensional finite element code DFLUX to investigate the heat flux distribution on the EAST divertor target plate and subsequently analyzed the thermal load characteristics under different heating scenarios~\cite{shi2018study}. Following the upgrade of EAST to a water-cooled W/Cu monoblock structure, two-dimensional models became inadequate for representing its complex geometric features. To address this limitation, Yang et al.~\cite{yang2020development} developed a three-dimensional finite element heat flux analysis method, enabling heat flux inversion based on infrared thermography measurements.
Physics-Informed Neural Networks (PINNs) integrate governing equations, boundary conditions, and observational data into the training process, thereby unifying data-driven learning with physical constraints. Hennigh et al.~\cite{hennigh2021nvidia} proposed SimNet, a multiphysics simulation framework; Cai et al.~\cite{cai2020heat} applied PINNs to heat transfer prediction under unknown boundary conditions; Peng et al.~\cite{peng2024multi} employed PINNs for thermal history prediction in additive manufacturing processes; Zhang et al.~\cite{zhang2022multi} developed M-PINN to solve heat conduction problems in multilayer materials; and Xu et al.~\cite{xu2023physics} utilized a physics-informed convolutional network to predict temperature and heat flux fields in porous media.
Neural operators have gained increasing attention in heat conduction problems because of their superior ability to generalize across different scenarios.
For forward problems, Zhang et al.~\cite{yuan2026transient} developed a transient temperature field prediction method based on the Fourier Neural Operator (FNO) and further extended it to unstructured meshes using graph neural operators, improving performance for complex geometries. For inverse problems, Li et al.~\cite{ding2026physics} proposed a physics-informed hierarchical neural operator framework that enables heat source identification and temperature field reconstruction under sparse measurements. For surrogate modeling, Wang et al.~\cite{cheng2025surrogate} combined a Fourier basis-enhanced DeepONet with uncertainty quantification techniques to achieve fast and reliable heat transfer prediction under flow fluctuations.

\section{Our Proposed Approach}  

\subsection{Overview} 

To accurately model the temperature field of the EAST divertor, we propose a Physics-aware Neural Operator Transformer (PNOT) framework, as shown in Fig.~\ref{fig::framework}. The framework integrates query point coordinates, global operating conditions, and heat flux information to learn the temperature field under physical constraints. By combining physics-aware attention with heat graph propagation, the proposed physical enhancement module captures both long-range physical dependencies and local heat-transfer characteristics. The network is trained with a hybrid loss of the mean squared error and graph Sobolev losses to improve both reconstruction accuracy and physical consistency.

\subsection{Problem Formulation}

We consider a PDE defined on a domain $\Omega \subset \mathbb{R}^{d}$. Let $A$ denote the input function space and $H$ the corresponding solution space. The input space $A$ includes boundary conditions, source functions, global system parameters, and other problem-specific inputs. We aim to learn a nonlinear operator $G$ that maps an input function to the corresponding PDE solution $G: A \rightarrow H$.

For the problem considered in this work, the goal is to learn a neural operator
\(
\mathcal{G}_{\theta}:\mathcal{C}\times\Omega\rightarrow T,
\)
which maps a distribution condition $\mathbf{c}\in\mathcal{C}$, where $\mathcal{C}\in \mathbb{R}^{d}$ denotes the distribution condition space, and an arbitrary spatial query coordinate $\mathbf{x}\in\Omega$ to the corresponding temperature value $T\in\mathbb{R}$.

\begin{equation}
\hat{T}(\mathbf{x})
=
\mathcal{G}_{\theta}(\mathbf{c},\mathbf{x}),
\qquad
\mathbf{x}\in\Omega.
\end{equation}
The learned operator is expected to generalize to distribution conditions that are not observed during training, thereby exhibiting strong out-of-distribution generalization capability.

\subsection{Network Architecture}

\subsubsection{Input Encoding}

Global operating conditions and heat flux boundary conditions jointly determine the temperature field, while the boundary heat flux acts as a distributed heat source that governs the local heat transfer process. To preserve the spatial structure of these physical conditions, we represent the global operating conditions and boundary heat flux as a structured token sequence rather than a conventional low-dimensional vector.

For the global conditions, we define 
$\mathbf{g}=\{\text{Sample Type}, K_{\mathrm{heat}}, y_i\}\in \mathbb{R}^d $
to denote the global operating conditions, where \textit{Sample Type} indicates the sampling category, 
$K_{\mathrm{heat}}$ represents the heat-source power, and $y_i$ denotes the control parameter. 
These global operating conditions are encoded into a \emph{Global Token} through the \emph{Global Branch}.

\begin{equation}
\mathbf{Z}_g=\phi_g(\mathbf{g})
\end{equation}
where $\phi_g(\cdot)$ is a learnable embedding network.

\begin{figure}
\centering
\includegraphics[width=0.95\textwidth]{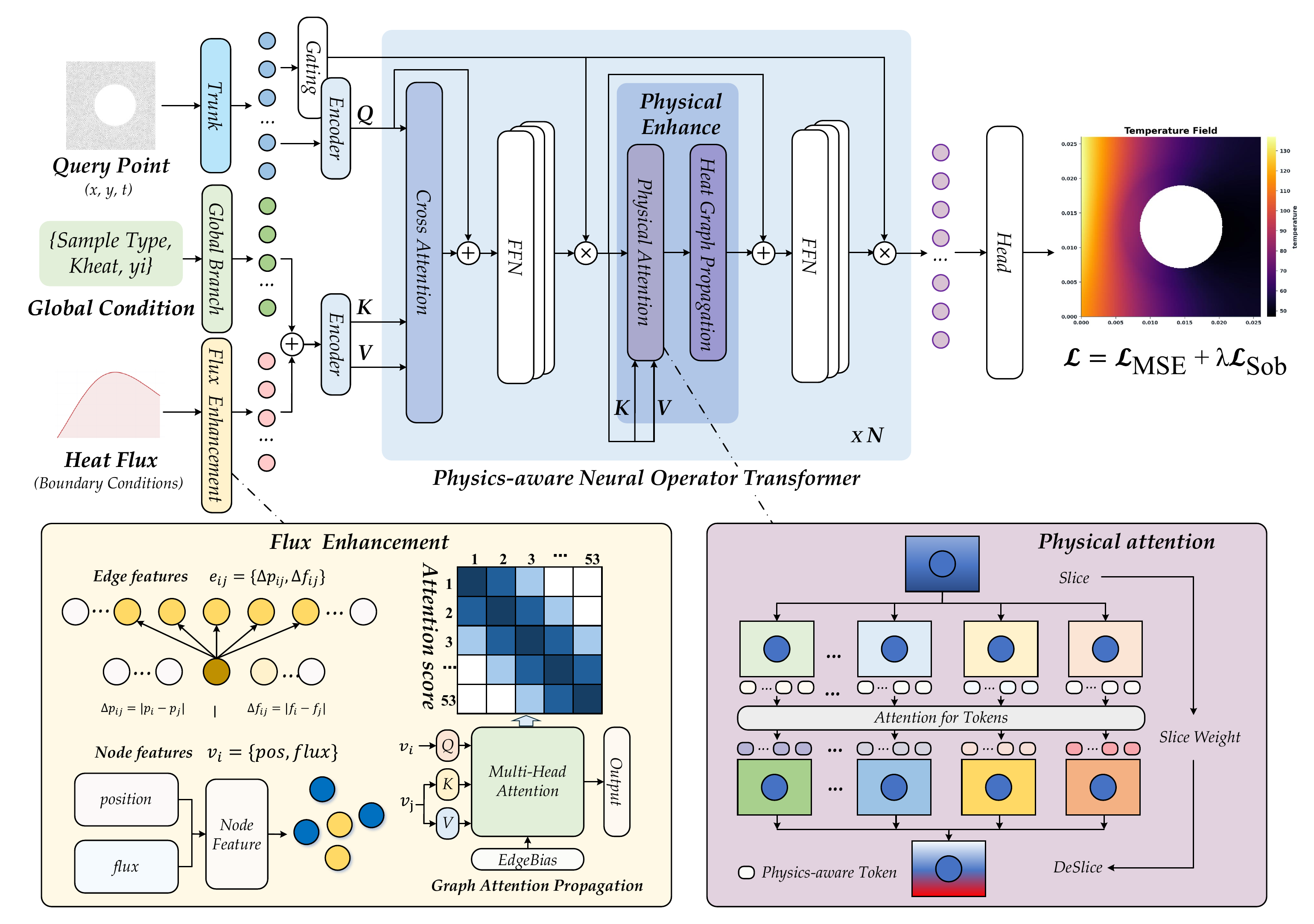} 
\caption{An overview of our proposed Physics-aware Neural Operator Transformer (PNOT) for temperature field reconstruction on EAST divertor.} 
\label{fig::framework}
\end{figure}

For heat flux boundary conditions, accurately modeling the spatial variation of boundary heat flux is essential for capturing the thermal response. As shown in Fig.~\ref{fig::framework}, we introduce a Flux Enhancement module to better capture local variations in boundary heat flux under thermal boundary conditions. Specifically, the boundary samples are first organized into a local graph according to their sampling order.
We define the graph as $\mathcal{G} = (\mathcal{V}, \mathcal{E})$, where each node $i \in \mathcal{V}$ corresponds to a boundary sampling point. The node feature is defined as:
\begin{equation}
\mathbf{v}_i = \{h_i, s_i\}
\end{equation}
where $h_i \in \mathbb{R}$ denotes the heat-flux value and $s_i \in \mathbb{R}$ denotes the normalized boundary coordinate.
Edges are constructed between neighboring nodes to model local interactions. For an edge $(i,j) \in \mathcal{E}$, the edge feature is defined as:
\begin{equation}
\mathbf{e}_{ij} = \{|s_i - s_j|,|h_i - h_j|\}
\end{equation}
For each node, query, key, and value features are obtained through learnable linear projections:
\begin{equation}
\mathbf{q}_i = W_Q \mathbf{x}_i,\quad
\mathbf{k}_i = W_K \mathbf{x}_i,\quad
\mathbf{v}_i = W_V \mathbf{x}_i,
\end{equation}
where $\mathbf{x}_i\in \mathbb{R}^{d} $ denotes the input node feature, obtained by applying an MLP to $\mathbf{v}_i$ and $W_Q$, $W_K$, and $W_V$ are learnable projection matrices.
To incorporate both geometric proximity and local heat-flux variation, we define an edge-aware bias term:
\begin{equation}
b_{ij} =
\phi_{\mathrm{edge}}
\left(
\mathbf{e}_{ij}
\right),
\end{equation}
where $\phi_{\mathrm{edge}}(\cdot)$ is a learnable edge encoder. 
The edge-aware attention propagation is then given by
\begin{equation}
\begin{aligned}
\mathbf{z}_i^{\mathrm{flux}} &= \sum_{j \in \mathcal{N}(i)} \operatorname{softmax}_{j \in \mathcal{N}(i)} \left( \frac{\mathbf{q}_i^{\top}\mathbf{k}_j}{\sqrt{d_h}} + b_{ij} \right) \mathbf{v}_j,  \\
\mathbf{Z}_{\mathrm{f}} &= [ \mathbf{z}_1^{\mathrm{flux}}, \mathbf{z}_2^{\mathrm{flux}}, \dots, \mathbf{z}_{53}^{\mathrm{flux}} ] \in \mathbb{R}^{53\times d}
\end{aligned}
\end{equation}
where $d_h$ is the feature dimension of each attention head and $\mathcal{N}(i)$ denotes the neighborhood of node $i$. The $\mathbf{Z}_{f}$ represents the enhanced \emph{Boundary Token} required in our framework. Finally, the Global Token and enhanced Boundary Tokens are concatenated to form the structured \emph{Branch Memory} $\mathbf{Z}_b=\left[\mathbf{Z}_g; \mathbf{Z}_{f}\right]$.

For the query points, 
We denote the set of spatio-temporal query coordinates as
$\mathcal{P}=\{(\mathbf{x}_i,t_i)\}_{i=1}^{N_q},$
where $\mathbf{x}_i \in \mathbb{R}^2$ represents the 2D spatial coordinate and $t_i \in \mathbb{R}$ represents the 1D temporal coordinate of the $i$-th query point.
These query points are encoded into \emph{Position Tokens} through the \emph{Trunk Branch}.

\begin{equation}
\mathbf{Z}_p=\phi_t(\mathbf{P})
\end{equation}
where $\phi_t(\cdot)$ is a learnable embedding network.


\begin{figure}[t]
\centering
\includegraphics[width=\textwidth]{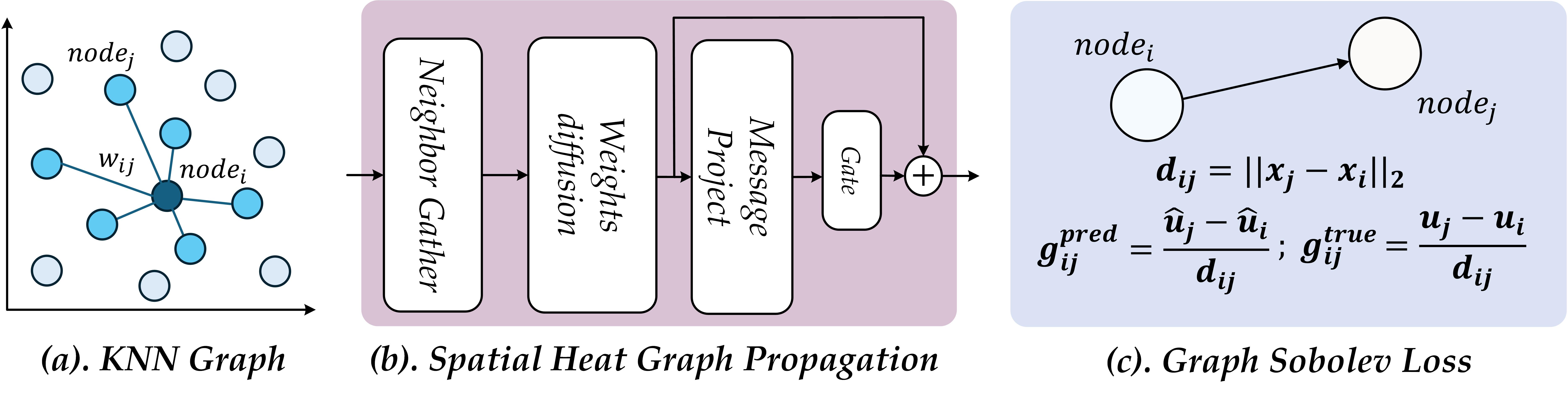} 
\caption{Overview of the proposed spatial heat graph propagation framework. (a) Construction of a K-nearest neighbor (KNN) graph from temperature sampling points. (b) Spatial heat graph propagation via neighbor gathering, weighted diffusion, message projection, and gated residual fusion. (c) Graph Sobolev loss for preserving local temperature gradients. }
\label{fig::framework-part2}
\end{figure}

\subsubsection{Physics-aware Neural Operator Transformer}
In heat conduction problems, query points are inherently correlated because the temperature field is governed by both global heat transfer and local diffusion. Therefore, independent point-wise prediction cannot fully capture spatial continuity or local variations. To address this, we explicitly model inter-query relationships in the backbone network, enabling the model to learn both global state correlations and local spatial couplings.

For the features associated with the query points, we first project them into the query representation using an encoder:
\begin{equation}
 \mathbf{Q} = \operatorname{Encoder}_{q}(\mathbf{Z}_{p}).   
\end{equation}
For the features derived from the global operating conditions and boundary heat flux, which constitute the Branch Memory $(\mathbf{Z}_{b})$, we encode them into the key and value representations:
\begin{equation}
\mathbf{K} = \operatorname{Encoder}_{k}(\mathbf{Z}_{b}), 
\qquad
\mathbf{V} = \operatorname{Encoder}_{v}(\mathbf{Z}_{b}).
\end{equation}
The query-point features and the conditional boundary features are then fused through a cross-attention mechanism:
\begin{equation}
\mathbf{Z}_{\mathrm{attn}}
=
\operatorname{softmax}
\left(
\frac{\mathbf{Q}\mathbf{K}^{\top}}{\sqrt{d_k}}
\right)
\mathbf{V}.   
\end{equation}
Through cross-attention, each query point can adaptively retrieve the most relevant global and boundary-condition information from the Branch Memory according to its spatial and physical context, leading to a condition-aware feature representation for subsequent temperature field reconstruction. The resulting feature is concatenated with the query vector of the query point and passed through an FNN for nonlinear transformation. The FNN output is then adaptively gated by the query point and fed into the physics-enhanced module.

For the physics-enhanced module, we jointly exploit global and local information to enhance the representation capability of physical features.

At the global level, the proposed model adopts the physical attention mechanism~\cite{wu2024transolver} to organize query representations. Given the encoded query features 
\(\mathbf{X}=\{\mathbf{x}_{i}\}_{i=1}^{N}\), a learnable soft-slicing operation assigns query points to \(G\) latent physical state slices:
\begin{equation}
\mathbf{W}=\operatorname{Softmax}(\operatorname{MLP}(\mathbf{X})),
\end{equation}
where \(\mathbf{W}\in\mathbb{R}^{N\times G}\) and \(w_{ig}\) denotes the assignment weight from the \(i\)-th query point to the \(g\)-th slice. The \(g\)-th state slice is computed as:
\begin{equation}
\mathbf{s}_{g}=\sum_{i=1}^{N} w_{ig}\mathbf{x}_{i}.
\end{equation}
Unlike spatial hard partitioning, this soft-slicing mechanism groups query points according to learned physical states, allowing distant points with similar physical characteristics to share the same slice. Self-attention is then performed among the slice representations:
\begin{equation}
\widetilde{\mathbf{S}}=\operatorname{Attention}(\mathbf{S}),
\end{equation}
and the updated slice features are projected back to each query point by:
\begin{equation}
\widetilde{\mathbf{x}}_{i}=\sum_{g=1}^{G}w_{ig}\widetilde{\mathbf{s}}_{g}.
\end{equation}
This slice--attention--deslice process captures global dependencies through \(G\) latent states rather than all \(N\) query points, reducing the attention cost while preserving global physical interactions.


As shown in Fig.~\ref{fig::framework-part2}, to capture local heat diffusion, we introduce a Heat Graph Propagation module based on the physical coordinates of query points. Specifically, the module constructs a spatial KNN graph and performs distance-weighted message propagation to model local heat transfer.
Given node feature $\mathbf{x}_i$ and coordinate $\mathbf{p}_i$, we first construct a $k$-nearest neighbor graph using
\begin{equation}
d_{ij}=\|\mathbf{p}_i-\mathbf{p}_j\|_2 .
\end{equation}
For each neighbor $j\in\mathcal{N}_k(i)$, the diffusion weight is computed as:
\begin{equation}
w_{ij}
=
\frac{\exp(-d_{ij}/\tau)}
{\sum_{m\in\mathcal{N}_k(i)}\exp(-d_{im}/\tau)},
\end{equation}
where $\tau$ is a learnable temperature parameter that controls the spatial diffusion range. The local diffusion message is then obtained from feature differences:
\begin{equation}
\Delta_i
=
\sum_{j\in\mathcal{N}_k(i)}
w_{ij}(\mathbf{x}_j-\mathbf{x}_i).
\end{equation}
This formulation resembles a graph Laplacian heat diffusion term and emphasizes local spatial variations. The message is further projected and injected into the token representation through a gated residual connection:
\begin{equation}
\mathbf{x}_i^{out}
=
\mathbf{x}_i
+
\operatorname{Dropout}
\left[
\tanh(\gamma)\cdot
\phi(\Delta_i)
\right],
\end{equation}
where $\phi(\cdot)$ denotes the message projection MLP, and $\gamma$ is a learnable scalar gate parameter. The term $\tanh(\gamma)$ adaptively controls the strength of the graph diffusion message. This module enables local thermal information to propagate along spatial neighborhoods, improving the modeling of fine-grained temperature distributions.

As shown in Fig.~\ref{fig::framework-part2} (c), to further preserve local spatial structures, we introduce a graph-based Sobolev regularization term. Consistent with the local graph propagation module, a $k$-nearest-neighbor graph is constructed over the queried spatial coordinates, and local spatial gradients are approximated by finite differences along graph edges. Let $\mathcal{N}_k(i)$ denote the set of $k$ nearest neighbors of query point $i$. For an edge $(i,j)$, the directional finite-difference gradient of the ground-truth field is defined as:
\begin{equation}
g_{i,j}
=
\frac{u_j-u_i}
{\|\mathbf{x}_j-\mathbf{x}_i\|_2},
\qquad
j\in\mathcal{N}_k(i),
\label{eq:graph_gradient}
\end{equation}
where $u_i$ and $u_j$ denote the ground-truth field values at query points $i$ and $j$, respectively. Here, $\mathbf{x}_i$ and $\mathbf{x}_j$ are the corresponding spatial coordinates.
Similarly, the directional finite-difference gradient of the predicted field is computed from the predicted values $\hat{u}_i$ and $\hat{u}_j$ along the same edge. The graph-based Sobolev loss is then formulated by penalizing the discrepancy between the predicted and ground-truth directional gradients:
\begin{equation}
\mathcal{L}_{\mathrm{Sob}}
=
\frac{1}{Nk}
\sum_{i=1}^{N}
\sum_{j\in\mathcal{N}_k(i)}
\left(
\frac{\hat{u}_{j}-\hat{u}_{i}}
{\|\mathbf{x}_{j}-\mathbf{x}_{i}\|_2}
-
\frac{u_{j}-u_{i}}
{\|\mathbf{x}_{j}-\mathbf{x}_{i}\|_2}
\right)^2.
\label{eq:sobolev_loss}
\end{equation}
where $\hat{u}_i$ and $\hat{u}_j$ denote the predicted field values at query points $i$ and $j$, respectively, and $k$ denotes the number of nearest neighbors connected to each query point.

\section{Experiments}

\subsection{Dataset and Evaluation Metric}

\begin{figure}
\centering
\includegraphics[width=\textwidth]{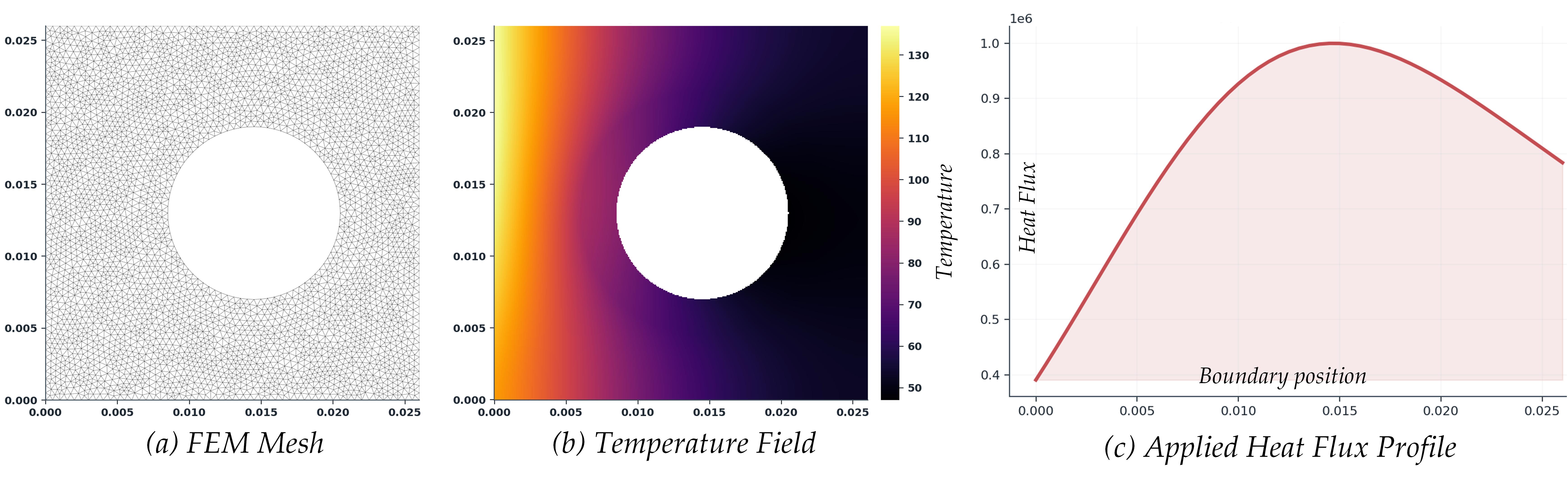} 
\caption{Representative samples from the heat conduction dataset with a heat source power of 1 MW: 
(a) finite element discretization mesh, (b) temperature distribution at the final simulation time, and (c) surface heat flux distribution.}
\label{fig::dataset}
\end{figure}

\noindent $\bullet$ \textbf{Dataset}:
As illustrated in Fig.~\ref{fig::dataset}, the dataset used in this study was generated through finite element simulations and is designed to characterize the evolution of transient temperature fields under different heat-source powers and boundary heat-flux conditions. The dataset contains ten heat-source power scenarios ranging from 1~MW to 10~MW. For each power level, 71 simulation samples are provided, resulting in a total of 710 samples. All simulations are performed on a unified finite element mesh consisting of 3,908 spatial nodes and 1,001 temporal snapshots. Each sample includes the heat-source power $K_{\mathrm{heat}}$, spatial coordinates $\mathbf{p}$, transient temperature field $\mathbf{u}$, as well as the corresponding boundary heat-flux distribution $\mathrm{Heat \ Flux}$ and sampling locations $x_{\mathrm{heat}}$. The heating boundary is discretized into 53 sampling points over a fixed interval of $[0,\,0.026]$~m. In addition, a control parameter $y_i$ is introduced to characterize different heat-flux distribution profiles. For each heat-source power level, 69 distinct $y_i$ configurations are considered, yielding a rich set of parameter combinations and thermal responses across the dataset. Consequently, the learning task can be formulated as an operator mapping from the physical parameters and boundary conditions, i.e., $(K_{\mathrm{heat}}, y_i, x_{\mathrm{heat}}, \mathrm{Heat \ Flux}, \mathbf{p})$, to the corresponding spatio-temporal temperature field $\mathbf{u}$.

\noindent $\bullet$ \textbf{Evaluation Metric}:
To evaluate the performance of the models, we take the relative L2 error (Rel L2), relative Mean Absolute Error (rMAE), relative Root Mean Square Error (rRMSE) and Mean Absolute Error (MAE). The metrics are formulated as: 
\begin{equation}
\mathrm{Rel \ L2}=\frac{1}{N}\sum_{i=1}^{N}\frac{\lVert \hat{u}(x_n, t_n) - u(x_n, t_n) \rVert_2}{\lVert u(x_n, t_n) \rVert_2 + 10^{-8}}, 
\end{equation}

\begin{equation} 
\text{rMAE}(\hat{u}) = \frac{\sum_{n=1}^N |\hat{u}(x_n, t_n) - u(x_n, t_n)|}{\sum_{n=1}^N |u(x_n, t_n)|},
\end{equation}
\begin{equation} 
\text{rRMSE}(\hat{u}) = \sqrt{\frac{\sum_{n=1}^N |\hat{u}(x_n, t_n) - u(x_n, t_n)|^2}{\sum_{n=1}^N |u(x_n, t_n)|^2}}, 
\end{equation}
\begin{equation} 
\text{MAE}(\hat{u}) = \frac{1}{N} \sum_{n=1}^N |\hat{u}(x_n, t_n) - u(x_n, t_n)|.
\end{equation}
where $ N $ is the number of test points, $ u(x, t) $ is the ground-truth solution, and $ \hat{u}(x, t) $ is the model's prediction.

\subsection{Implementation Details} 
The network consists of three stacked PNOT blocks. The boundary relation encoder uses a neighborhood size of 6, while the spatial heat graph is constructed with $k=8$. The dataset is split into training, validation, and test sets with a ratio of 8:1:1. During training, one temporal snapshot and 1024 spatial query points are randomly sampled from each field sample, whereas all spatial nodes are used for inference. The model is trained for 300 epochs using the AdamW optimizer with an initial learning rate of $1\times10^{-3}$, a weight decay of $5\times10^{-6}$, and a cosine annealing learning rate scheduler with a minimum learning rate of $1\times10^{-5}$. The batch size is set to 4. The training objective combines the mean squared error (MSE) loss with a graph-based Sobolev regularization term weighted by $\lambda=0.01$. All experiments are conducted on a single NVIDIA GeForce RTX 3090 GPU. More details can be found in our source code.

\subsection{Comparison with State-of-the-art Models}

As shown in Table~\ref{tab:comparison}, we compare the proposed PNOT with a broad range of representative operator learning methods, including DeepONet~\cite{lu2021learning}, wavelet-based MWT~\cite{gupta2021multiwavelet}, Fourier-based Geo-FNO~\cite{li2023fourier} and F-FNO~\cite{tran2021factorized}, geometry-aware models such as GNOT~\cite{hao2023gnot}, OTNO~\cite{li2025geometric}, and RIGNO~\cite{mousavi2026rigno}, Transformer-based Transolver~\cite{wu2024transolver}, and pretrained operator model DPOT~\cite{hao2024dpot}. In addition, U-NO~\cite{rahman2022u}, LSM~\cite{wu2023solving}, and TNO~\cite{calvello2025continuum} are also included for a comprehensive comparison.

The results in Table~\ref{tab:comparison} show that PNOT consistently achieves the lowest reconstruction errors across all considered evaluation metrics, including Rel~L2, rRMSE, rMAE, and MAE. Compared with existing operator learning models, PNOT demonstrates more accurate reconstruction of the divertor heat-flux temperature field. 

\begin{table*}[t]
\caption{Comparison of different operator learning models.}
\label{tab:comparison}
\resizebox{\columnwidth}{!}{ 
\centering
\small
\renewcommand{\arraystretch}{1.15}
\setlength{\tabcolsep}{8pt}
\begin{tabular}{l|l|cccc}
\toprule
\textbf{Model} &\textbf{Publish}  & \textbf{Rel\ L2} $\downarrow$ & \textbf{rRMSE} $\downarrow$ & \textbf{rMAE} $\downarrow$ & \textbf{MAE} $\downarrow$  \\
\midrule
DeepONet~\cite{lu2021learning}          & Nat. Mach. Intell. 2021  &0.0056   &0.0055  &0.0043  &0.7142    \\
MWT~\cite{gupta2021multiwavelet}        & NeurIPS 2021  &0.0614 & 0.0560 & 0.0527 & 9.8221    \\
F-FNO~\cite{tran2021factorized}         & arXiv 2021  & 0.4656 & 0.5004 & 0.4752 & 99.4731  \\
U-NO~\cite{rahman2022u}                 & arXiv 2022  &0.0037 & 0.0023 & 0.0022 & 0.4652   \\
Geo-FNO~\cite{li2023fourier}            &JMLR 2023  &0.6234 & 0.3850 & 0.3849 & 80.5688    \\
LSM~\cite{wu2023solving}                &arXiv 2023  &0.0970 & 0.0807 & 0.0726 & 15.2168  \\
GNOT~\cite{hao2023gnot}                 &ICML 2023  &0.0022 & 0.0016 & 0.0014 & 0.3114   \\
Transolver~\cite{wu2024transolver}      &ICML 2024  &0.0058 & 0.0056 & 0.0050 & 0.9510\\
DPOT~\cite{hao2024dpot}                 &ICML 2024  & 0.0055 & 0.0038 & 0.0036 & 0.7595\\
OTNO~\cite{li2025geometric}             &JMLR 2025  &0.0078 & 0.0071 & 0.0048 & 1.0193    \\
TNO~\cite{calvello2025continuum}        &JMLR 2025  &0.0066 & 0.0048 & 0.0044 & 0.9384   \\
RIGNO~\cite{mousavi2026rigno}           &NeurIPS 2026   &0.0027 & 0.0018 & 0.0017 & 0.3707   \\
\midrule 
PNOT (Ours)                             &-   &  \textbf{0.0008} & \textbf{0.0006} & \textbf{0.0006} & \textbf{0.1224}    \\
\bottomrule
\end{tabular}
} 
\end{table*}



\subsection{Component Analysis}

To evaluate the effectiveness of each proposed component, we conduct ablation experiments on the finite element simulation temperature field dataset. The results are reported in Table~\ref{tab:ablation}.

\noindent $\bullet$  \textbf{Boundary Enhancement (BT).}
The Boundary Enhancement module explicitly models the spatial relationships among boundary heat fluxes, allowing the boundary representations to better capture the influence of boundary conditions on the interior temperature field. Incorporating BT reduces the relative error from \textbf{0.0021} to \textbf{0.0015}, demonstrating that enhanced boundary representations provide more informative physical cues for temperature field prediction.

\noindent $\bullet$ \textbf{Sobolev Loss (SL).}
The proposed Sobolev loss introduces gradient supervision to encourage consistency between the predicted and ground-truth temperature fields in the local neighborhood. Compared with using only point-wise regression, this constraint enables the model to learn smoother and more physically consistent temperature distributions. Consequently, the relative error is further reduced from \textbf{0.0015} to \textbf{0.0011}, confirming the effectiveness of incorporating local physical constraints during optimization.

\noindent $\bullet$ \textbf{Physical Module (PAHG).}
The physical module integrates Physics Attention and Heat Graph Propagation to capture complementary global physical dependencies and local heat diffusion characteristics. By jointly modeling these two aspects, the model is able to learn a more comprehensive physical representation of the temperature field. As a result, the relative error is further reduced to \textbf{0.0008}, achieving the best performance among all configurations.

The results demonstrate that each proposed component consistently contributes to performance improvement. Boundary Enhancement strengthens the utilization of boundary information, Sobolev Loss enhances local physical consistency, and the physical module further exploits complementary global and local physical correlations. When all components are combined, the proposed model achieves the lowest relative error of \textbf{0.0008}, corresponding to a \textbf{61.9\%} reduction compared with the baseline. These results verify that the proposed components are complementary and jointly improve both prediction accuracy and physical consistency.

\begin{table}[t]
\centering
\caption{Ablation study of PNOT. "BT" denotes boundary tokens, "SL" denotes Sobolev loss, "PAHG" denotes physics attention (PA) and heat graph propagation (HG).}
\label{tab:ablation}
\setlength{\tabcolsep}{8pt}
\renewcommand{\arraystretch}{1.15}
\begin{tabular}{l|ccc|cc}
\toprule
\textbf{Method} & \textbf{BT} & \textbf{SL} & \textbf{PAHG}   & \textbf{Rel\ L2} $\downarrow$  \\
\midrule
Baseline &  &  &  & 0.0021 \\
+ Boundary Enhanced & \checkmark &  &  &0.0015\\
+ Sobolev Loss & \checkmark & \checkmark&  &0.0011 \\
+ Physical Module & \checkmark & \checkmark & \checkmark &\textbf{0.0008} \\
\bottomrule
\end{tabular}
\end{table}

\subsection{Ablation Study} 

\noindent $\bullet$ \textbf{Analysis on tradeoff parameter $\lambda$.~} 
As shown in Table~\ref{tab:lambda}, the tradeoff parameter $\lambda$ has a clear impact on model performance. 
As $\lambda$ increases from $1e^{-4}$ to $1e^{-2}$, all metrics consistently improve, indicating that a proper weight helps balance the optimization objectives. 
The best performance is obtained at $\lambda=1e^{-2}$, achieving Rel L2, rRMSE, rMAE, and MAE of $0.000868$, $0.000612$, $0.000602$, and $0.122425$, respectively. 
However, when $\lambda$ is further increased to $1e^{-1}$, the performance drops, suggesting that an excessively large weight may disturb the optimization balance. 
Thus, we set $\lambda=1e^{-2}$ in our experiments.

\begin{table}[t]
\centering
\caption{Analysis on the tradeoff parameter $\lambda$.}
\label{tab:lambda}
\begin{tabular}{c|cccc}
\toprule
$\ \ \ \lambda \ \ \ $ & \ \  \textbf{Rel\ L2 } \ \ & \ \ \textbf{rRMSE} \ \  &\ \  \textbf{rMAE} \ \ &\ \  \textbf{MAE}\ \   \\
\midrule
  1e-4 & 0.0018 & 0.0013 & 0.0012 & 0.2541 \\
  1e-3 & 0.0013 & 0.0008 & 0.0008 & 0.1865 \\
  \textbf{1e-2} & \textbf{0.0008} & \textbf{0.0006} & \textbf{0.0006} & \textbf{0.1224}  \\
  1e-1 & 0.0017 & 0.0013 & 0.0012 & 0.2497 \\
\bottomrule
\end{tabular}
\end{table}

\noindent $\bullet$ \textbf{Analysis on number of PNOT blocks.~}
As shown in Table~\ref{tab:pnot_blocks}, the number of PNOT blocks affects the model performance. 
When the number of blocks increases from $1$ to $3$, all metrics consistently improve, indicating that stacking more PNOT blocks enhances the representation capability. 
The best performance is achieved with $3$ blocks, achieving the best performance across all evaluation metrics, including Rel L2, rRMSE, rMAE, and MAE. 
However, further increasing the number of blocks to $4$ or $5$ leads to performance degradation, suggesting that excessive blocks may introduce optimization difficulty or redundancy. 
Therefore, we use $3$ PNOT blocks in our experiments.

\begin{table}[t]
\centering
\caption{Analysis on the number of PNOT blocks.}
\label{tab:pnot_blocks}
\begin{tabular}{c|cccc}
\toprule
\ \ \ Blocks \ \ \ & \ \  \textbf{Rel\ L2 } \ \ & \ \ \textbf{rRMSE} \ \  &\ \  \textbf{rMAE} \ \ &\ \  \textbf{MAE}\ \   \\
\midrule
  1  & 0.0011 & 0.0007 & 0.0007 & 0.1574 \\
  2  & 0.0010 & 0.0006 & 0.0006 & 0.1413 \\
   \textbf{3} & \textbf{0.0008} & \textbf{0.0006} & \textbf{0.0006} & \textbf{0.1224}  \\
  4 & 0.0009 & 0.0006 & 0.0006 & 0.1356 \\
  5 & 0.0010 & 0.0007 & 0.0007 & 0.1424 \\
\bottomrule
\end{tabular}
\end{table}

\begin{figure}[t]
    \centering
    \includegraphics[width=1\linewidth]{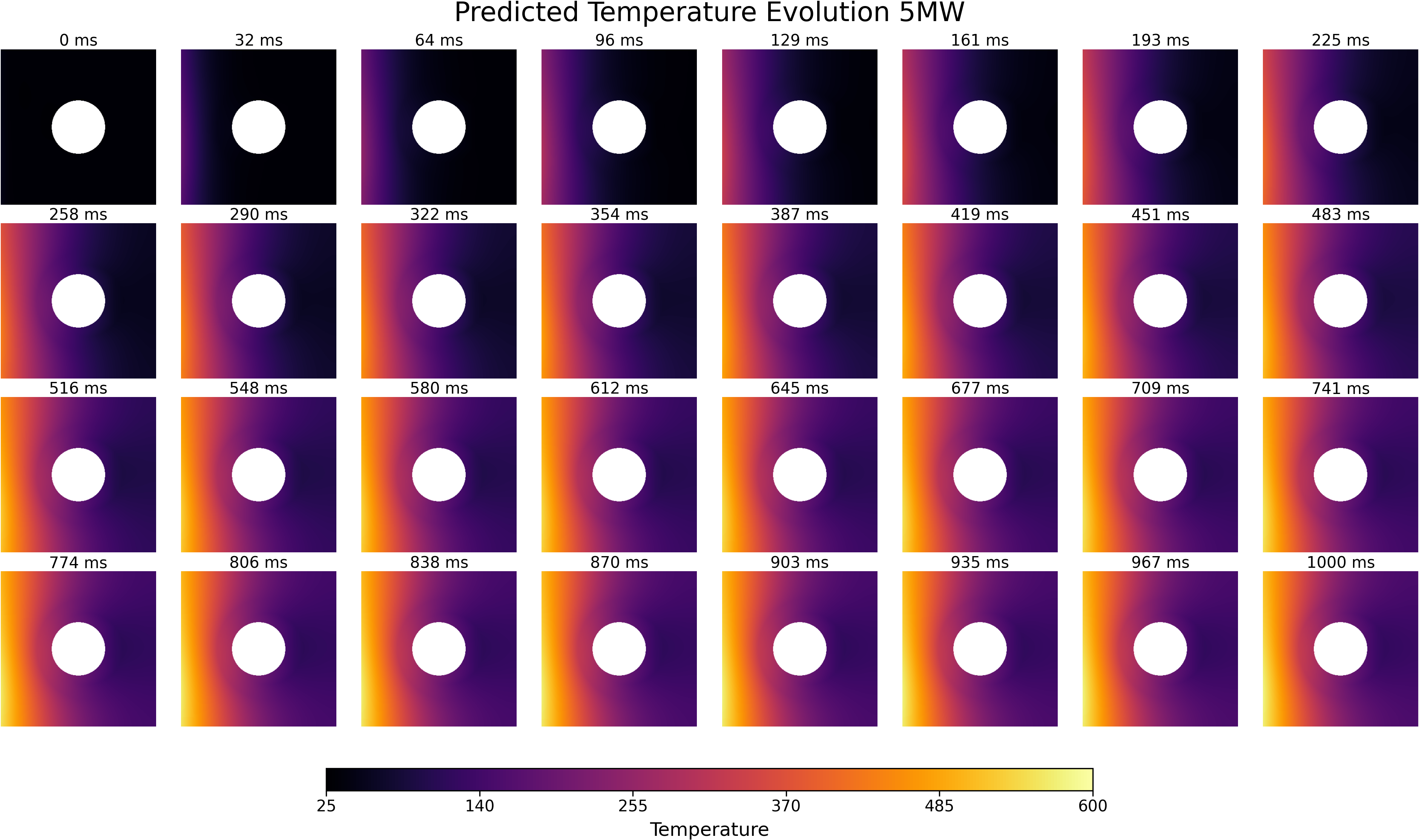}
    \caption{Temporal evolution of temperature under the 5 MW operating condition}
    \label{fig:temperature-time}
\end{figure}

\noindent $\bullet$ \textbf{Influences of the parameter K on KNN Graphs.~} 
As shown in Table~\ref{tab:knn_k}, the choice of parameter $K$ in KNN graphs significantly influences the model performance. 
The model achieves the best results when $K=8$. 
When $K$ is smaller, such as $2$ or $4$, the graph may capture insufficient neighborhood information, resulting in inferior performance. 
In contrast, increasing $K$ to $16$ or $32$ degrades all metrics, suggesting that overly dense neighborhoods may introduce irrelevant connections and noise. 
Therefore, we set $K=8$ in our experiments.

\begin{table}[t]
\centering
\caption{Influences of the parameter $K$ on KNN graphs.}
\label{tab:knn_k}
\begin{tabular}{c|cccc}
\toprule
$\ \ \ K \ \ \ $  & \ \  \textbf{Rel\ L2 } \ \ & \ \ \textbf{rRMSE} \ \  &\ \  \textbf{rMAE} \ \ &\ \  \textbf{MAE}\ \   \\
\midrule
  2  & 0.0009 & 0.0006 & 0.0006 & 0.1291 \\
  4  & 0.0009 & 0.0006 & 0.0006 & 0.1389 \\
  \textbf{8}  &\textbf{0.0008}  & \textbf{0.0006} & \textbf{0.0006} &\textbf{0.1224} \\
  16 & 0.0010 & 0.0007 & 0.0007 & 0.1496 \\
  32 & 0.0011 & 0.0008 & 0.0008 & 0.1633 \\
\bottomrule
\end{tabular}
\end{table}


\subsection{Visualization} 
As shown in Fig.~\ref{fig:temperature-time}, the temporal evolution of the reconstructed temperature field under the 5~MW (Out-of-Distribution) operating condition. As the heating process proceeds, the divertor temperature field gradually develops and eventually reaches a quasi-steady state. During the initial stage (0--1000~ms), the temperature remains relatively low, with a noticeable increase only near the heat-loaded boundary. As time progresses, heat diffuses into the interior, leading to the formation of a stable temperature gradient after approximately 500~ms. The proposed model accurately captures the transient-to-steady evolution of the temperature field, producing smooth and physically consistent predictions without noticeable numerical oscillations. 

\begin{figure}
    \centering
    \includegraphics[width=1\linewidth]{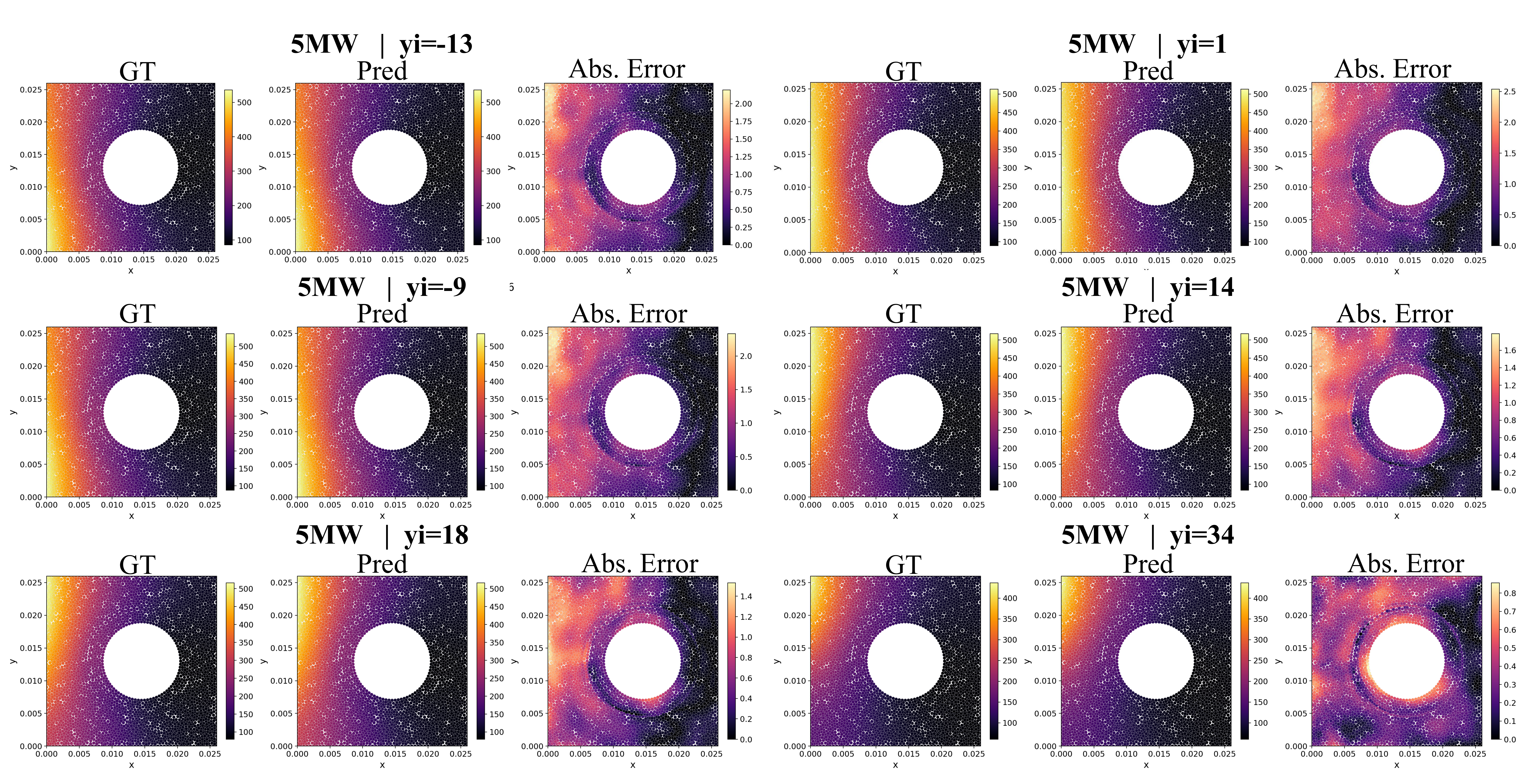}
    \caption{Comparison of the reconstructed temperature fields by the proposed model and the finite element method under different $y_i$ conditions at 5\,MW.}
    \label{fig:abs_error}
\end{figure}

To further evaluate the spatial reconstruction performance, Fig.~\ref{fig:abs_error} compares the reconstructed temperature fields with the finite element solutions under different $y_i$ conditions at 5~MW. The predicted results show excellent agreement with the FEM solutions in terms of the overall temperature distribution, temperature gradients, and the locations of high-temperature regions. The absolute error remains small over most of the domain and is mainly concentrated near the circular boundary and regions with large temperature gradients, where the reconstruction task is inherently more challenging. Nevertheless, the model maintains consistently high reconstruction accuracy across different $y_i$ conditions, demonstrating its robustness and strong spatial generalization capability.

\subsection{Limitation Analysis} 

Although this study is trained and evaluated on finite-element–generated data under multiple heat flux conditions, demonstrating a certain degree of generalization capability, several limitations remain. First, the present work is based on two-dimensional modeling and numerical validation, whereas the heat flux distribution in a divertor is inherently a three-dimensional complex phenomenon; thus, the 2D approximation introduces non-negligible discrepancies from the real physical process. Second, compared with the operating environment of actual fusion devices, finite-element simulation data remain relatively idealized and do not fully capture the strongly coupled multi-physics effects and uncertainties present in real conditions, leading to potential gaps in real-world applicability. Finally, the proposed model is specifically developed for the EAST divertor system rather than as a general-purpose model, and its generalization ability across different operating scenarios and devices has yet to be systematically validated. Overall, further investigation is required to improve its robustness under more complex operating conditions and its applicability to practical engineering scenarios.

\section{Conclusion}
We propose PNOT, a physics-aware neural operator transformer for fast reconstruction of transient temperature fields in tungsten monoblock divertors. The proposed framework integrates structured boundary encoding, physics-guided attention, heat conduction modeling, and a Sobolev-constrained loss to effectively learn the nonlinear mapping from boundary heat flux to temperature field evolution. Experiments on the EAST finite-element dataset demonstrate that PNOT consistently outperforms existing neural operator methods in reconstruction accuracy while exhibiting superior stability and generalization under complex heat flux conditions, validating its effectiveness for modeling spatially coupled heat conduction. Nevertheless, the current study has been validated primarily on finite-element simulation data. Future work will extend the framework to real experimental measurements and more challenging operating conditions, enabling a more comprehensive assessment of its real-world applicability and potential for engineering deployment.


%

\bibliography{ref}
\bibliographystyle{splncs04}

%





\end{document}